\documentclass[11pt,a4paper]{article}
\usepackage[hyperref]{acl2019}
\usepackage{times}
\usepackage{latexsym}
\usepackage{graphicx}
\usepackage{amsmath}
\usepackage{bm}
\usepackage{booktabs}
\usepackage[ruled,linesnumbered]{algorithm2e}
\usepackage{algpseudocode}
\usepackage{mdwlist}
\usepackage{multirow}
\usepackage{url}
\usepackage{caption}
\usepackage{subfigure}
\usepackage{graphicx}

\aclfinalcopy 
\def\aclpaperid{2187} 

\title{Course Concept Expansion in MOOCs with External Knowledge and Interactive Game}

\newcommand*{\affaddr}[1]{#1}

\newcommand*{\email}[1]{\texttt{#1}}

\author{Jifan Yu$^{1,2}$, Chenyu Wang$^{3}$, Gan Luo$^{1,2}$ ,Lei Hou$^{1,2}$\thanks{corresponding author}, Juanzi Li$^{1,2}$, Jie Tang$^{1,2}$, Zhiyuan Liu$^{1,2}$\\
\affaddr{$^{1}$Dept. of Computer SCi.\& Tech., Tsinghua University, China 100084}\\
\affaddr{$^{2}$KIRC, Institute for Artificial Intelligence, Tsinghua University, China 100084}\\
\affaddr{$^{3}$Shenyuan Honors College, Beihang University, China 100083}\\
\email{\{yujf18@mails.,luog18@mails.\}tsinghua.edu.cn}\\
\email{\{houlei@,lijuanzi@,jietang@,liuzy@\}tsinghua.edu.cn}\\
\email{wangchenyu@buaa.edu.cn}\\
}

\date{}

\begin{document}
\maketitle
\begin{abstract}
    
    As Massive Open Online Courses (MOOCs) become increasingly popular, it is promising to automatically provide extracurricular knowledge for MOOC users. Suffering from semantic drifts and lack of knowledge guidance, existing methods can not effectively expand course concepts in complex MOOC environments. In this paper, we first build a novel boundary during searching for new concepts via external knowledge base and then utilize heterogeneous features to verify the high-quality results. In addition, to involve human efforts in our model, we design an interactive optimization mechanism based on a game. 
    Our experiments on the four datasets from Coursera and XuetangX show that the proposed method achieves significant improvements(+0.19 by MAP) over existing methods. The source code and datasets\footnote{Source Codes \& Datasets: \url{https://github.com/thukg/concept-expansion-kg}} have been published.
\end{abstract}

\section{Introduction}





\begin{figure}[ht]
    \centering
    \setlength{\abovecaptionskip}{0.2cm}
    \setlength{\belowcaptionskip}{-0.3cm}
    \includegraphics[width=0.95\linewidth]{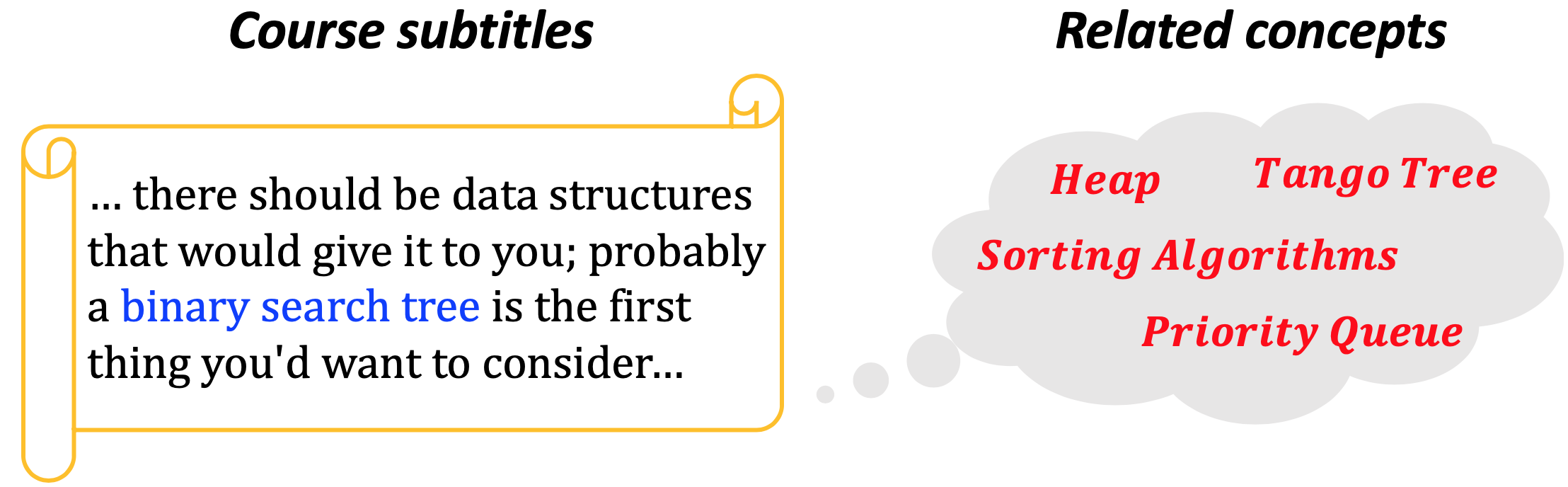}
    \caption{An example of ``out-of-teaching" concepts in the course ``\emph{Data Structure and Algorithm}''.}
    \label{exampleInMOOCs}
\end{figure}

Self-determination theory was first formally proposed by Deci and Ryan in \cite{deci1991motivation}, suggesting that educators should support students in autonomously discovering and learning course-related knowledge. In fact, in addition to the concepts taught in course, many related concepts are also worthy of learning. Figure \ref{exampleInMOOCs} shows a real example from Coursera in \emph{Data Structure} course. When the concept \emph{Binary Search Tree} is taught, some other concepts, including its similar structures (\emph{Heap}), applications (\emph{Sorting} and \emph{Priority Queue}) and advanced researches (\emph{Tango Tree}\footnote{It is an online binary search tree that achieves an $O(\log \log n)$ competitive ratio. \cite{demaine2007dynamic}}) also benefit students for further course understanding. However, these concepts are not available without specifical mention, especially in the era of Massive Open Online Courses (MOOCs).
In MOOCs, teachers need to keep a moderate length of the course to face with thousands of students with various backgrounds \cite{jordan2015massive}, making it infeasible to manually pick out these helpful concepts. Therefore, there is a clear need to automatically identify course-related concepts, so that the students can easily acquire additional knowledge and achieve better educational outcomes.

Although extracting course concepts from teaching materials \cite{kay2002automatic} or course subtitles \cite{pan2017course} has attracted several attempts, the research in finding the concepts absent in course materials, which we call \textbf{Course Concept Expansion}, has not been explored. Despite abundant work on related topics, including concept expansion or set expansion \cite{wang2007language,wang2015concept,10.1007/978-3-319-73305-0_12},
it is far from sufficient to directly apply these methods in the MOOC environments due to the following challenges.

First, unlike the set expansion for a clear general category (e.g., country), course concepts are often the combinations of multiple categories, which is easy to cause semantic drift \cite{curran2007minimising} during exploring in different domains (such as Structures: \emph{Heap}, \emph{Binary Tree} and \emph{Algorithms}: \emph{Divide and Conquer}, \emph{Greedy Algorithm}). Second, the features for manifesting course-related concepts are heterogeneous. As shown in Figure \ref{exampleInMOOCs}, we regard \emph{Heap} as a course concept due to its similar structure while \emph{Binary Search Tree} is a prerequisite concept of \emph{Tango Tree}. Thus mere context information is not enough for effective expansion.
Third, as an application-oriented task, it is beneficial to involve human interactions. How to properly leverage the feedback from MOOC users to obtain a better performance for concept expansion remains a challenging issue.


To address the above problems, we propose a three-stage course concept expansion model. Inspired by the idea of concept space \cite{hori1997concept}, we first build an accurate boundary for a given course to alleviate the semantic drift during candidate concept generation from an external knowledge base.
Then we transform the expansion into a binary classification problem as previous positive unlabeled learning methods for set expansion \cite{li2010distributional,wang2017sentiment}. Three types of features are proposed to incorporate heterogeneous information into classifier to identify high-quality concepts among candidates.
Finally, we design a lightweight but attractive top-student game to subtly collect MOOC users' feedback and iteratively optimize the expansion results.
\begin{figure}[ht]
    \setlength{\abovecaptionskip}{0.2cm}
    \setlength{\belowcaptionskip}{-0.3cm}
    \centering
    \includegraphics[width=0.9\linewidth]{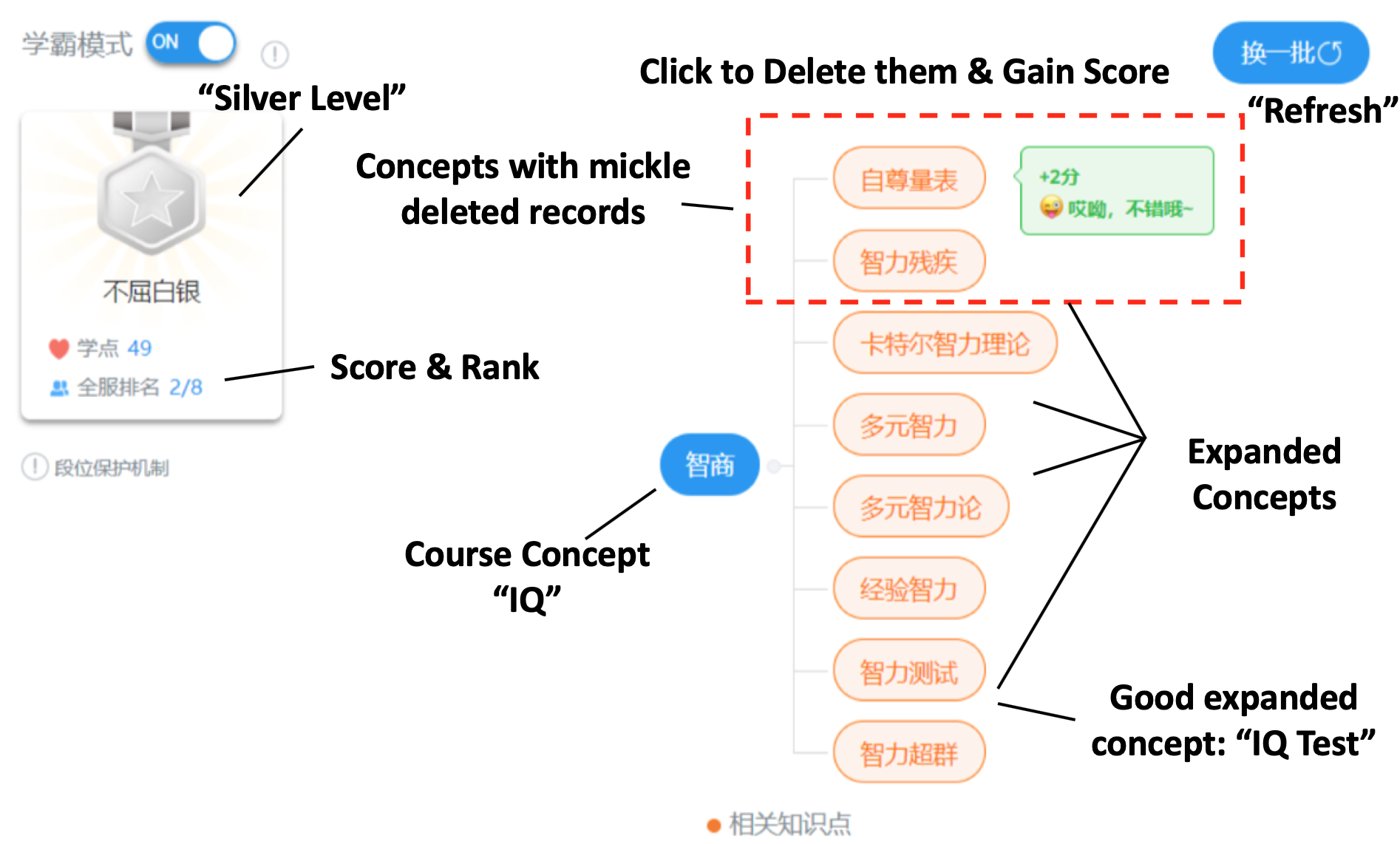}
    \caption{Top-Student Game in course "Introduction to Psychology" of XuetangX.}
    \label{game design}
\end{figure}

For evaluation, we compare the proposed method with 4 representative set expansion methods on real courses from Coursera\footnote{\url{https://www.coursera.org/}} and XuetangX\footnote{\url{http://www.xuetangx.com/}}, and further conduct online evaluation in the game mechanism.

\textbf{Contributions.} Our contributions include: a) systematically investigate the problem of course concept expansion in MOOCs; b) proposal of an effective three-stage model for course concept expansion using an external knowledge base and interactive game; c) four benchmark datasets using real courses from Coursera and XuetangX.






\section{Problem Formulation}

In this section, we first give some necessary definitions and then formulate the problem of course concept expansion.

\textbf{A Course corpus} is composed by $n$ courses in the same subject area, denoted as $\mathcal{D} = \left \{ \mathcal{C}_{j} \right \}_{j=1}^n$, where $\mathcal{C}_{j}$ is one course. We assume that course $\mathcal{C}_{j} = \left \{ v_{ij} \right \}_{i=1}^{m_{j}}$ consists of $m_{j}$ course videos, where $v_{ij}$ stands for the $i$-th video. Following \cite{pan2017course}, we define \textbf{Course Concepts} are the subjects taught in the course denoted as $\mathcal{M} = \left \{ c_{i} \right \}_{i=1}^{\left | \mathcal{M} \right |}$.

Existing work can extract course concepts $\mathcal{M}$ from course corpus $\mathcal{D}$, but $\mathcal{D}$ could inevitably miss some important course concepts (as illustrated in Figure \ref{exampleInMOOCs}). Therefore, there is a clear need to expand the course concepts beyond $\mathcal{M}$ using external resources. In this paper, we focus on the expansion using external knowledge bases.

\textbf{Knowledge Base} is formally defined as $\mathcal{KB} = \left ( E,R \right )$, where $E = \left \{ e_{i} \right \}_{i=1}^{\left | E \right |}$ represents all concepts, $R = \left \{ r_{i} \right \}_{i=1}^{\left | R \right |}$ represents the relationships between concepts, and $(e_{i},r_{j},e_{k})$ is a triple in $\mathcal{KB}$ meaning $e_{i}$ has relationship $r_{j}$ with $e_{k}$.

\textbf{Course Concept Expansion Using Knowledge Base in MOOCs} is formally defined as follows. Given the course concepts $\mathcal{M}$ and knowledge base $\mathcal{KB}$, Course Concept Expansion returns a ranked list of expanded concepts $E_c\subset E$, and outputs $s_{i}$ for $e_{i}\in E_c$ to indicate its likelihood to be an expanded concept of $\mathcal{D}$.

\section{Method}

To appropriately expand course concepts in MOOCs, we need to address three crucial problems. 1. How to alleviate semantic drift? 2. How to employ heterogeneous information to identify high-quality expanded concepts? 3. How to properly involve human efforts to optimize the expansion result?
In this section, we introduce our novel course concept expansion model in three stages.
\begin{figure}[ht]
    \centering
    \includegraphics[width=0.9\linewidth]{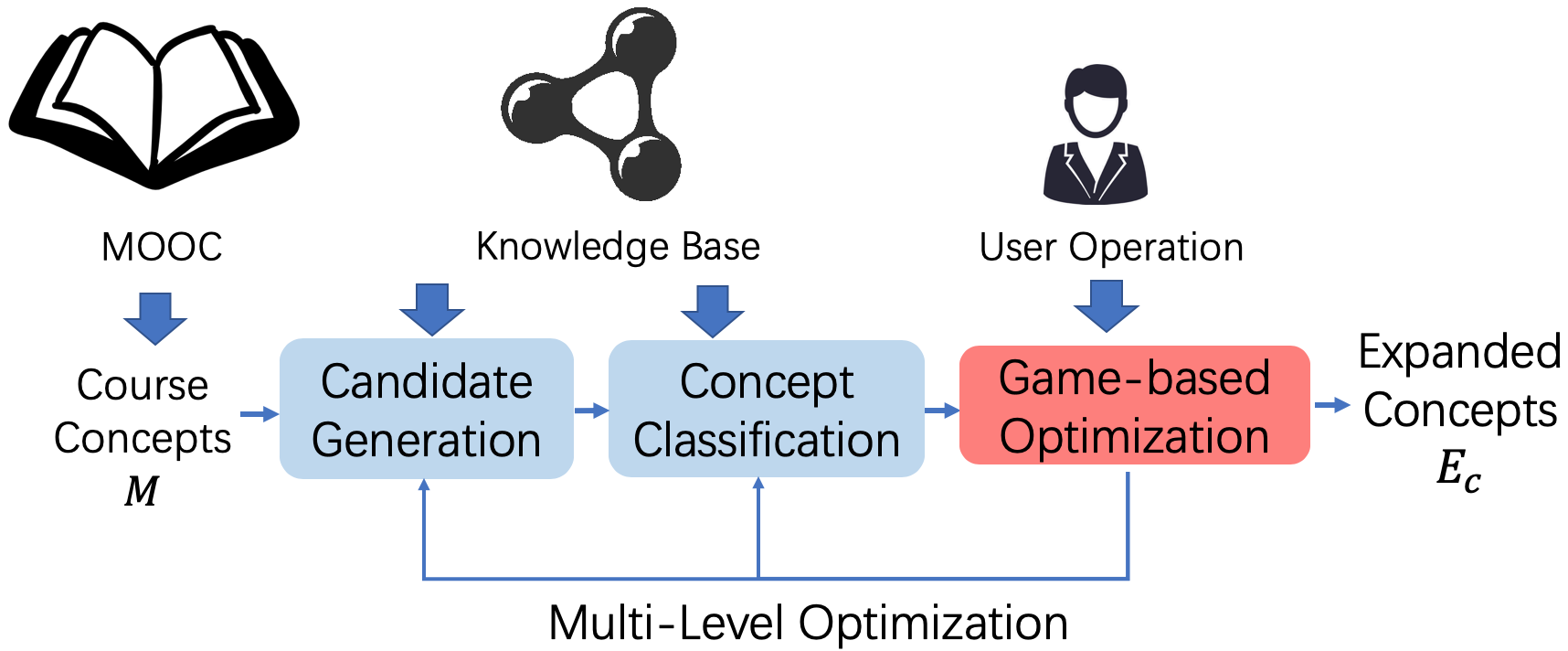}
    \caption{Framework of course concept expansion.}
    \label{framew}
\end{figure}

\begin{figure*}[ht]
    \centering
    \includegraphics[width=1.0\linewidth]{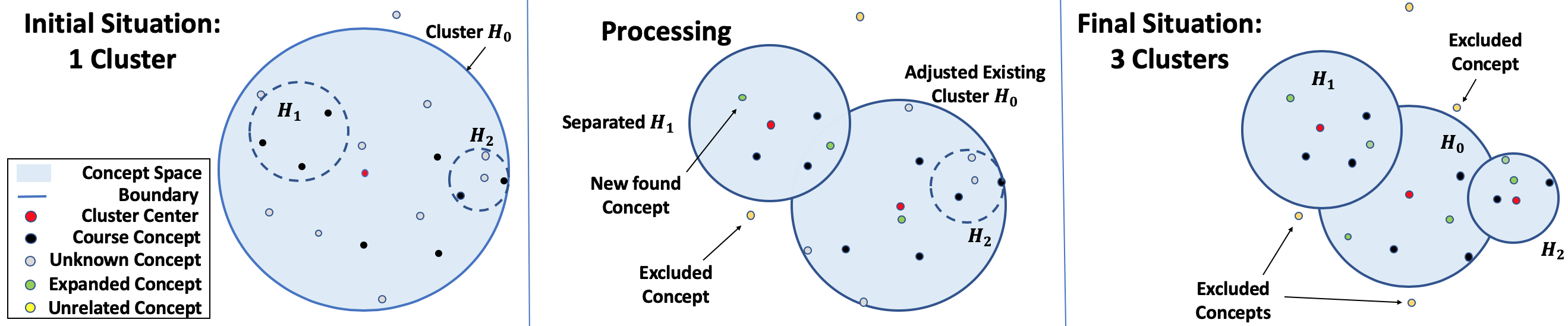}
    \caption{The concept space boundary is fitted in process of searching candidates. }
    \label{Concept-Space}
\end{figure*}

\textbf{(1)Candidate Generation}: To reduce semantic drift, a dynamic boundary is set during our searching for new concepts in $\mathcal{KB}$. We only admit the concepts within the boundary as candidates.

\textbf{(2)Concept Classification}: To leverage heterogeneous information in expansion, we propose three types of novel features to build a classifier, identifying the high-quality expansion concepts among candidates and rerank the result list.

\textbf{(3)Game-based Optimization}: To involve human efforts, we creatively design an interactive online game named top-student game which has been applied in a real MOOC platform to collect user feedback and cyclically optimize the expansion process at multiple levels.


\subsection{Candidate Generation}

In this section, we present an unsupervised embedding based algorithm that iteratively generates ranked concept candidates from an external knowledge base $\mathcal{KB}$. In particular, we build a boundary to avoid semantic drift based on the following concept space assumption.
\newtheorem{assumption}{Assumption}[section]
\begin{assumption}
A course is a concept space which contains one or more concept clusters.
\end{assumption}

The idea of concept space was proposed and applied in digital teaching and ontology engineering \cite{hori1997concept,cassidy2006using}. We extend the assumption by considering concepts' polycentric clustering pattern (e.g., Data Structure course's concepts mainly gather in several clusters, such as graph algorithms, binary trees, etc.). As shown in Figure \ref{Concept-Space}, we initialize the course concepts in one loose cluster and separate the gathering ones into new clusters during candidate generation. An explicit boundary of course is dynamically formed by its clusters to avoid semantic drift.


Given the course corpus $\mathcal{D}$ and knowledge base $\mathcal{KB}=(E, R)$, we first utilize the method in \cite{pan2017course,pan2017prerequisite} to extract course concepts $\mathcal{M}$. Note that we remove the extracted results which do not exist in $E$, i.e., $\mathcal{M}\subset E$, to facilitate following candidate generation. We use bold-face letters  to denote the embeddings of the corresponding terms (i.e., $\mathbf{c_{i}}$ is the embedding of $c_{i}$).

Before introducing the algorithm details, we first define the concept cluster as follows.
\newtheorem{definition}{Definition}[section]
\begin{definition}
\label{def:cc}
A concept cluster $H$ is formed by several semantically related course concepts $\{ c_{i} \}_{i=1,\ldots,|H|}$, and is formally represented as a hypersphere $(\mathbf{o}, \gamma)$ with $\mathbf{o}$ and $\gamma$ denoting its center and radius respectively. Mathematically,
\begin{displaymath}
\small
\begin{aligned}
\mathbf{o} &= \sum\nolimits_{c_i\in H}\mathbf{c_i}/|H|\\
\gamma &= \max\nolimits_{c_i\in H} edis(\mathbf{o},\mathbf{c_i})
\end{aligned}
\end{displaymath}
where $edis(\cdot ,\cdot)$ returns the Euclidean distance between the input vectors.
\end{definition}
Note that the center $\mathbf{o}$ may be a ``virtual" concept, i.e., it does not correspond to any known concept in $\mathcal{M}$ or $E$. To facilitate the generation process, we introduce a special subset $S_{H}\subset H$ that includes a fixed size $\tau$ ($\tau$ is experimentally set to 8) of representative concepts as seeds. We always select the ``actual" concepts nearest to the center $\mathbf{o}$, which means that $S_{H}$ might change dynamically during the generation process. The candidate generation algorithm contains two phases: initialization and searching.

\noindent \textbf{Initialization}: We initialize a concept cluster $H_0$ with all the concepts in $\mathcal{M}$ (as shown in Figure \ref{Concept-Space}), calculate its center $\mathbf{o}_{H_0}$ and radius $\gamma_{H_0}$, and select the representative subset $S_{H_0}$. Then following a predefined order\footnote{The course concepts are extracted with the method proposed in \citet{pan2017course}, which also assigns a confidence score for each concept. Here we sort the extracted concepts by the confidence score in descending order.}, we adapt the single-pass online clustering \cite{guha2003clustering} to group the concepts into potential clusters. The clustering algorithm sequentially processes the concepts, one at a time, and grows clusters incrementally. A concept $c_i$ is absorbed by a previously-generated cluster $H_i$ if its Euclidean distance to a concept in $H_i$ is below a predefined threshold\footnote{In experiment, it is set to the minimal distance between representative concepts and the center of $H_0$, i.e., $\min\nolimits_{c\in S_{H_0}} edis(\mathbf{o}_{H_0},\mathbf{c})$}; otherwise, the concept is treated a new potential cluster. Finally, we successfully partition the course concepts $\mathcal{M}$ into $L$ potential clusters $H_1,H_2,\ldots, H_{L}$.

\noindent \textbf{Searching}: For each concept $c_{ij}$ in a potential cluster $H_i$, we search its directly-connected concepts in knowledge base $\mathcal{KB}$, e.g., $(c_{ij}, r, e)$ with $e\in E$ and $r\in R$, and use the distance between $e$ and $c_{ij}$ to determine whether to merge it into $H_i$. Similar to the single-pass clustering, and merge it into $H_i$ if $edis(e, c_{ij}) < \min\nolimits_{c\in S_{H_i}} edis(\mathbf{o}_{H_i},\mathbf{c})$. During the process, we separate a cluster from $H_0$ whenever its size reaches $\tau$ (i.e., it is big enough to select representative concepts), update $H_0$ (including $\mathbf{o}_{H_0}$, $\gamma_{H_0}$ and $S_{H_0}$) and calculate its own parameters. For those potential clusters whose size are less than $\tau$, we use $\mathbf{o}_{H_0}$ and and $S_{H_0}$ to make the above decision. When $e$ is merged into $H_i$, we define the following confidence score $s_{e}$ to measure its likelihood to be a course concept,
\begin{equation}
\small
\setlength{\abovedisplayskip}{1pt}
\setlength{\belowdisplayskip}{1pt}
\begin{aligned}
s_{e} = cos(\mathbf{e},\mathbf{c}_{ij})+\sum\limits_{c_{ik}\in H_i} cos(\mathbf{c}_{ik},\mathbf{c}_{ij})\times cos(\mathbf{e},\mathbf{c}_{ik})
\end{aligned}
\end{equation}
where $cos(\cdot ,\cdot)$ returns the cosine similarity of the input vectors.

After the expansion for all concepts in $\mathcal{M}$, we obtain the expanded concept set $E_c^1\in E$. Then we sort $E_c^1$ by $s_{e}$ in descending order and iteratively repeat the search phase for $E_c^1$ to obtain $E_c^2$. The algorithm stops when no concepts in $E$ could be merged. Finally, we achieve $E_c=\bigcup\nolimits_{i}E_c^i$ and sort it by $s_{e}$ in descending order as candidate set.

It's worth noting that each final candidate $e\in E_c$ is directly or indirectly related to a course concept $c\in \mathcal{M}$. $e$ and $c$ are connected by a search path $c\rightarrow r_1\rightarrow e_1\rightarrow \ldots \rightarrow e$, where $e_i\in E$ and $r_i\in R$ are concept and relation in $\mathcal{KB}$, and we record such path (denoted as $path(e)$) to hold more semantics. The whole process is summarized in Algorithm~\ref{alg:cg}. Specifically, due to the huge number of operations for finding nearest concepts, we use K-D Tree\footnote{K-D tree is a useful data structure for nearest neighbor searches (Wikipedia).} to store concept vectors, which greatly improves the time efficiency.

\subsection{Concept Classification}


To integrate heterogeneous information in expansion, we propose three features from various sources and rerank the candidates after classification. As a binary classification problem, all existing classifiers can be applied, and we experimentally select XGboost \cite{chen2016xgboost}. In this section, we introduce the three types of features and how we partially rerank the candidates.

\noindent \textbf{Confidence Score}. In accordance to our assumption, the confidence score $s_{e}$ represents the degree of remoteness between the candidate concept $e$ and the concept cluster $H$ to which it belongs. Thus we select it as the first feature to capture a candidate's basic relevance to the course.

\noindent \textbf{Search Path Encoding}. During candidate generation, the search paths insinuate the semantic relations between course concepts. Taking ``\emph{Floyd Algorithm}" as an example, its search path, ``BFS $\rightarrow$ \emph{InstanceOf} $\rightarrow$ Graph Algorithms $\rightarrow$ \emph{Instance} $\rightarrow$ Floyd Algorithm", indicates that ``Floyd Algorithm is a sibling of course concept ``BFS". To make effective use of this semantic information, we employ an RNN encoder-decoder neural network \cite{cho2014learning} to encode $path(e)$ for candidate $e$. Specifically, we train the neural network to take $path(e)$ as input and output the same sequence. Thus, we can obtain a fixed-length vector representation of $path(e)$ from the final hidden state of the RNN encoder.
\SetKwRepeat{Do}{do}{while}
\begin{algorithm}[t]
\small
\caption{Candidate Generation}
\label{alg:cg}
\KwIn{$\mathcal{M}$, $\mathcal{KB}=(E,R)$, $\tau$}
\KwOut{$E_c$}

Sort $\mathcal{M}$, initialize $H_0$ and further partition into $H_1,H_2,\ldots, H_{L}$\\
$t=0; E_c^0= \mathcal{M}$\\
\Do {$E_c^t\neq \emptyset$}{
    $E_c^{t+1}= \emptyset$ \\
    \For {$c_{ij} \in H_i \subset E_c^t$ has related concept $e\in E$}{
        \If {$edis(e, c_{ij}) < \min\nolimits_{c\in S_{H_i}} edis(\mathbf{o}_{H_i},\mathbf{c})$}{
            $E_c^{t+1}=E_c^{t+1}\cup \{e\}$\\
            Merge $e$ into $H_i$, calculate $s_{e}$, record $path(e)$ and update $H_i$\\
            \If {$|H_i|\geq \tau$}{
                Separate $H_i$ from $H_0$ and update $H_0$
            }
        }
        $E_c=E_c\cup E_c^{t+1}$
    }
    Sort $E_c^{t+1}$ by $s_e$ and $t += 1$
}
Sort $E_c$ by $s_e$
\end{algorithm}

\noindent \textbf{Prerequisite Features}. The course concepts also have an unique relationship called \emph{Prerequisite} \cite{margolis1999concepts}. Prerequisite concept pair $\left ( A,B \right )$ means if someone wants to study A, he/she is better to understand B in advance (e.g., \emph{Binary Tree} is a prerequisite concept to \emph{Black-Red Tree}), which indicates how concepts in the course are connected.
There are a few previous efforts to extract prerequisite relations from Wikipedia \cite{talukdar2012crowdsourced,liang2015measuring}, textbooks \cite{yosef2011aida,wang2016using} and MOOCs \cite{pan2017prerequisite}. In this paper, we select five features from \cite{pan2017prerequisite} that only rely on the course text, and $Pv(a, b)$ is the combination of these five features reflecting the prerequisite likelihood of $a$ to $b$. Since these features can only measure the relationship between the two concepts that exist in the course, we calculate the prerequisite feature of $e$ using its search root phrase $c_{i}$ as follows:
\begin{equation}
\small
\setlength{\abovedisplayskip}{2pt}
\setlength{\belowdisplayskip}{2pt}
\begin{aligned}
Pf(e) = \frac{cos\left \langle \mathbf{e},\mathbf{c_{i}} \right \rangle \ast \sum_{c_{j}\in \mathcal{M}} Pv(c_{i},c_{j})}{\left | \mathcal{M} \right |}
\end{aligned}
\end{equation}

\noindent \textbf{Partial Reranking}. After feature extraction and classification, each candidate is labeled with a tag P (positive) or N (negative). Then we partially adjust the rankings in low-confidence interval to improve the recall. We define a threshold $\alpha \in \left [ 0,1 \right ]$ to control the reranking range and adjust rankings after $\alpha \ast \left | E_c \right |$. Specifically, we sort positive and negative results separately according to their confidence score $s$ and then place the positive results before negative ones. Eventually a reranked expansion list is achieved.






\subsection{Game-based Optimization}

As an application-oriented task, it is beneficial to properly introduce human efforts to monitor and optimize our expansion model. However, the design of this human-model interaction faces several challenges. For human, we need to ensure the quality and sufficiency of their feedback. For the model, we need to effectively and fully utilize the provided data in model optimization.

Thanks to the multimedia and web platform of MOOCs~\cite{volery2000critical}, we are able to attractively collect feedback from students with diverse backgrounds using an online game, which far exceeds traditional modes in amount. In this section, we introduce the game design and how the collected feedback optimize our model.


\noindent \textbf{Game Design.} We design our feedback collector, a game named ``Top-Student" by considering its attractiveness and the quality of collected data.

To make game attractive, we place the game under each video $v_{ij}$ of course $\mathcal{D}$, whose basic idea is that the player gains the score and competes with other students of the course by deleting low-quality expanded concepts. It allows users to quickly start with a simple click-to-delete operation. Further, we only show expansion results that are most relevant to the concepts in the video $v_{ij}$, which increases the affinity of the users who just finished the video. Those design facilitates a wider collection of feedback.

Figure \ref{game design} is the game layout in the course of ``Introduction to Psychology". The blue concept ``IQ" is a course concept $c$ in video, while the orange concepts are its relevant expanded concepts $E_c$. Users delete low-quality concepts among orange ones and gain different scores while we always record the total deletion times of each expanded concept $e_{i}$'s (denoted as $del(e_{i})$).

To ensure data quality, we avoid users' irresponsible deletion by employing a group-vote scoring mechanism. Specifically, when a user deletes the expanded concept $e_{i}$, he/she gets a score $Q = del(e_{i})/\max_{e_{j}\in E_c} del(e_{j})-\frac{1}{2}$, i.e., every user operation corresponds to a score $ S \in \left [ -\frac{1}{2},\frac{1}{2} \right ]$ (In real application, the value is enlarged to $\left [-5,5 \right ])$ ) based on all existing deletion data, which means that irresponsible operations subject to a penalty.

We set up the game by calculating and storing the expanded concept $e\in E_c$ with highest cosine similarity for each $c_{i}$ as inputs. Finally we get total deletion $del(e)$ of each $e$ as outputs. The Top-Student Game has been applied in several courses at one of largest Chinese MOOC websites, XuetangX and collected over 10,000 records as of this writing.





\noindent \textbf{Multi-level Optimization.} The user feedback affects both candidate generation in Section 3.1 and concept classification in Section 3.2 to perform a multi-level optimization.

We first define and calculate the deletion ratio of a candidate $e_i$ as $Dr(e_{i}) = del(e_{i})/\max_{e_{j}\in E_c}(del(e_{j}))$. The value reflects the acceptance of $e_i$ comparing with the other expansion results. Then we present the optimization at two levels.


\textbf{Candidate Generation}: The confidence score $s_{e}$ in Equation 1 is updated, reducing the likelihood from its directly related concept according to its deletion ratio $Dr(e)$.
\begin{equation}
\small
\setlength{\abovedisplayskip}{1pt}
\setlength{\belowdisplayskip}{1pt}
\begin{aligned}
s_{e} = &cos(\mathbf{e},\mathbf{c}_{ij})\times (1-Dr(e)) \\&+\sum\limits_{c_{ik}\in H_i} cos(\mathbf{c}_{ik},\mathbf{c}_{ij})\times cos(\mathbf{e},\mathbf{c}_{ik})
\end{aligned}
\end{equation}

\textbf{Concept Classification}: We regard deletion ratio $dr(e)$ as a new feature to incorporate user insights into our classifier.

In this way, user feedback is applied to each process of the model, and new results generated after optimization are also periodically entered into the game to collect feedback again. Finally it iterates over and gets high-quality expansion results.

\section{Experimental Evaluation}


\subsection{Datasets}


Since there is no publicly available dataset for course concept expansion in MOOCs, we use two different domains of Chinese and English courses: ``Data Structure and Algorithm" and ``Introduction to Psychology" to construct four datasets through a three-stage process.


First, for each domain, we select its most relevant English courses from Coursera and Chinese courses from XuetangX, e.g., for EN-DSA, we select 3 courses (The three courses: Algorithms (Princeton), Algorithms (Stanford), Data Structure and Algorithm (UC San Diego)) of 3 universities and obtain a total of 449 videos. Then, we use the method of Pan \shortcite{pan2017course} to extract the course concepts and manually select the high-quality ones as the course concepts $\mathcal{M}$. Finally, we take XLORE~\cite{jin2018xlore2} as $\mathcal{KB}$ to search for related course concepts and manually labeled the reasonable expansion results. For each domain, we collect 100,000 related concepts and record their search path to train the encoder in Section 3.2. But the large amount requires arduous human labeling work, thus we only pick \textbf{800} expanded concepts with the highest average cosine similarity to the course concepts to label. For each concept, two human annotators majoring in the corresponding domain are asked to label them as ``0: Not related" or ``1: Related " based on their own knowledge. Thus, each dataset is doubly annotated, and pearson \emph{correlation} coefficient is applied to assess inter-annotator agreement. A candidate is labeled as a related concept only if the two annotators are in agreement. For each dataset, we split it into training (400), validation (200) and test set (200).

Table \ref{Dataset} presents the detailed statistics, where \emph{\#courses}, \emph{\#videos}, \emph{$|\mathcal{M}|$}, \emph{1-Label} and \emph{0-Label} are the number of courses, videos, course concepts, positive and negative labels. We can only obtain \emph{\#deletions} from game for Chinese datasets.

\begin{table}[htbp]
\small
  \centering
    \begin{tabular}{r|cc|cc} \toprule
\multirow{2}{*}{}& \multicolumn{2}{c|}{\textbf{DSA}} & \multicolumn{2}{c}{\textbf{PSY}} \\ \cline{2-5}
          & ZH & EN & ZH & EN\\ \hline
    \emph{\#courses} & 1     & 3 & 1     & 1 \\ \hline
    \emph{\#videos} & 490   & 465 & 57    & 478 \\ \hline
    \emph{$|\mathcal{M}|$}   & 305   & 201 & 575   & 470 \\ \hline
    \emph{1-Label} & 398   & 232 & 237   & 246 \\ \hline
    \emph{0-Label} & 402   & 568 & 563   & 554 \\ \hline
    \emph{correlation} & 0.696 & 0.734 & 0.712 & 0.681 \\ \hline
    \emph{\#deletions} & 6939  & -     & 4920   & - \\ \bottomrule
    \end{tabular}%
\caption{Datasets Statistics}
\label{Dataset}
\end{table}%

\subsection{Experiment Settings}


\textbf{Basic Setting.} We choose GloVe \cite{pennington2014glove} as our English word embedding, \cite{li2018analogical} as our Chinese word embedding. We follow the same process of \cite{cho2014learning} to train the path encoder and \cite{pan2017prerequisite} to get prerequisite features for classifier in Section 3.2.

\noindent \textbf{Baseline Methods.} We compare our models (simple candidate generation results denoted as MOOC) with four typical methods which employ different similarity metrics. MOOC-C means our model with Classification and MOOC-CG means the whole model added game optimization.

\noindent $\bullet$ \textbf{PR} Graph based method: We build the candidates and course concepts into a graph. When the similarity between two concepts exceeds a threshold $\tau_{PR}$, there is a link between them. The PageRank score of each candidate is finally used for sorting. A most famous method employing pagerank is SEAL \cite{wang2007language}.

\noindent $\bullet$ \textbf{SEISA} SEISA\cite{he2011seisa} is an entity set expansion system developed by Microsoft after SEAL and outperforms traditional graph-based methods by an original unsupervised similarity metric. We implement its Dynamic Thresholding algorithm to sort expanded concepts.

\noindent $\bullet$ \textbf{EBM} Embedding based method mainly utilizes context information to examine the similarity between expanded concepts and seeds like \cite{mamou2018term}. For each expanded concept $e$, we calculate the pairwise cosine similarity with course concepts $\mathcal{M}$ in word2vec and use the average as golden standard to rank the expanded concept list.

\noindent $\bullet$ \textbf{PUL} PU learning is a semi-supervised learning model regarding set expansion as a binary classification task. We employ the same setting as \cite{wang2017sentiment} to classify and sort concepts.






\noindent \textbf{Evaluation Metrics.} Our objective is to generate a ranked list of expanded concepts. Thus, to evaluate the ranking result, we use the \textbf{Mean Average Precision}(MAP) as our evaluation metric, which is the preferred metric in information retrieval for evaluating ranked lists.

\subsection{Overall Evaluation}

\begin{table*}[t!]
\centering
\begin{tabular}{c|ccc|ccc}
\toprule
 & \textbf{ZH-PSY} & \textbf{ZH-DSA} & \textbf{ZH-Avg} & \textbf{EN-PSY} & \textbf{EN-DSA} & \textbf{EN-Avg} \\ \hline
PR & 0.849  & 0.519  & 0.684  & 0.833  & 0.822  & 0.827 \\
SEISA & 0.877  & 0.448  & 0.663  & 0.814  & 0.890  & 0.852  \\
EBM & 0.875  & 0.421  & 0.648  & 0.777  & 0.858  & 0.817  \\
PUL & 0.855  & 0.680  & 0.768  & 0.734  & 0.795  & 0.765  \\\hline
MOOC & 0.894  & 0.785  & 0.839  & 0.781  & 0.933  & 0.857  \\
MOOC-C & 0.939  & 0.804  & 0.872  & {\bfseries 0.922}  & {\bfseries 0.976}  & {\bfseries 0.949} \\
MOOC-CG & {\bfseries 0.954}  & {\bfseries 0.819}  & {\bfseries 0.886}  & -  & -  & - \\ \bottomrule
\end{tabular}
\caption{MAP of different methods on datasets. ($S_{H}=8, \alpha=0.4$).}
\label{map}
\end{table*}

Table \ref{map} summarizes the comparing results of different methods on all datasets, and -avg means the average MAP of datasets in same language. We find that our method outperforms existing methods across all 4 datasets. The improvements are all statistically significant tested with bootstrap re-sampling with 95\% confidence. For example, our whole model surpasses PageRank based method and SEISA by about 0.10 on the average of English courses. Further, we observe the performance in following aspects:

\noindent \textbf{For different datasets}, our methods stably perform at a competitive level while existing methods fluctuate fiercely. All methods maintain a better result in English than Chinese. Especially in ZH-DSA, existing methods meet a sharp decline at over 0.17. To find out the reason, we calculate the average pairwise similarity between the extracted concepts $\mathcal{M}$ in each dataset. Results show that ZH-DSA contains the most scattered course concepts at a pairwise distance of 0.60 (ZH-PSY, EN-PSY, EN-DSA at 0.49, 0.50, 0.36). But our expansion achieves a fine result in ZH-DSA at 0.785, indicating it effectively relieves the semantic drift after candidate generation.

\noindent \textbf{For different components of our methods.} The pure candidate generation (MOOC) mainly improves the performance by obvious promotion in ZH-DSA. The governing improvement exists after classification (at an average over 0.90), verifying the effectiveness of heterogeneous features we proposed. Moreover, the game-based optimization further improves the performance (+0.14), which proves the power of human efforts and our feedback optimization.

\subsection{Result Analysis}
\begin{figure}
\centering
\subfigure[The MAP Curve of $\tau$ (when $\alpha = 0.4$)] { \label{fig:a}
\includegraphics[width=0.46\columnwidth]{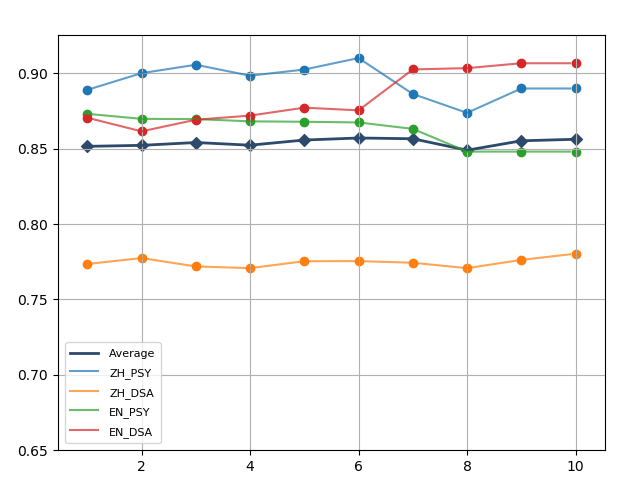}
}
\subfigure[The MAP Curve of $\alpha$ (when $\tau = 8$)] { \label{fig:b}
\includegraphics[width=0.46\columnwidth]{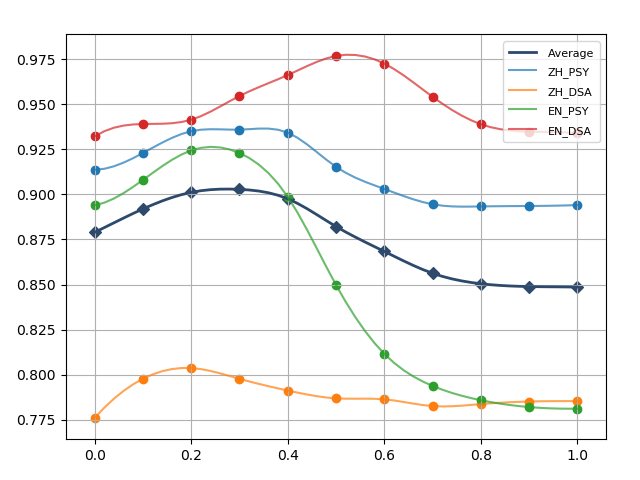}
}
\caption{Parameter analysis.}
\label{fig}
\end{figure}

\noindent \textbf{The size of seed set $\tau$.} The seed set size $\tau$ controls how many concepts of $E_c$ and $\mathcal{M}$ should be employed to calculate confidence score. We adjust $\tau$ from 1 to 10 and explore the influence of $\tau$ on Candidate Generation. Figure \ref{fig:a} shows the MAP transmutation. Despite different setting of $\tau$, our model maintain a preeminent competitive performance at an average MAP of 0.85 for English courses and 0.81 for Chinese courses.

\noindent \textbf{Feature Contribution Analysis.} To evaluate the features proposed in Section 3.2 and 3.3,with highest average F1-score at 0.94 as our classifier and run our approach 4 times on the 2 Chinese datasets, with one different feature deleted in each test. Table \ref{contribution} records the changes of $P$, $R$ and $F_{1}$ for each setting. According to the decrement of F1-scores, we find that all the proposed features are indispensable in classification. Especially, we observe that search path encode $Pe$ plays the most important role, decreasing most F1-score by 7.96\%. Besides, user deletion $Dr$ from game also outstandingly increases the precision of classifier by 8.21\%.
\begin{table}[h]
\centering
\small
\begin{tabular}{c|c|c|c}
\toprule
Ignored Feature & $P$ & $R$ & $F_{1}$ \\ \hline
$s_e$ & -2.65\% & -1.73\% & -1.83\% \\
$Pe$ & -3.97\% & {\bfseries -10.6\%} & {\bfseries -7.96\%} \\
$Pf$ & -7.39\% & -1.55\% & -3.78\% \\
$Dr$ & {\bfseries -8.21\%} & -3.10\% & -4.93\% \\
\bottomrule
\end{tabular}
\caption{Contribution analysis of different features. $s_e$, Pe, Pf, Dr are respectively confidence score, search path encoding, prerequisite features and deletion ratio.}
\label{contribution}
\end{table}

\noindent \textbf{The ranking threshold $\alpha$.} The ranking threshold $\alpha $ is the parameter controls how much ratio of results in Section 3.1 should be adjusted by classification. As we increase $\alpha$, less candidates will be adjusted, which weakens the role of the classifier. In Figure \ref{fig:b}, we set $\alpha $ from 0 to 1 and find that the performance reaches a peak at an average 0.3 of $\alpha$. The results demonstrate extra information provided by classifier effectively lifts the recall in low-confidence area (latter 70\% in average).

\subsection{Online Evaluation}


In particular, our model uses a gamified form to build a human interface. This interactive design not only optimizes the model, but can also be used to evaluate the effectiveness of the model in practical teaching applications.

By collecting deletion data of the expanded concepts, we can peep whether our expansion results are really helpful to the MOOC users. In order to quantitatively observe the user's feedback on the model, we propose \textbf{Average Correction Rate}($C_{r}$) as a game-to-model evaluation metric. This metric is the ratio of the number of times the user deletes the top $n$ concepts to the total deletions, and is formally denoted as $C_{r} = \sum_{i=1}^{n} del(e_{i}) / \sum_{j=1}^{\left | E_c \right |} del(e_{j})$. We set $n$ to 10, 50, 100, and the $C_{r}$ of each methods are listed in Table \ref{ACR}. Higher $C_{r}$ indicates less users think the expansion results are helpful. From this perspective, our method is the most helpful to them. For $C_r$@10, PUL performs a slight advantage at 0.002. However, once the range of observation is broadened to 50 or 100, our method shows an obvious ascendancy. Besides, after adding classifier and game, user satisfaction further increases, reducing the $C_r$ by 0.004 and 0.049 at $C_r$@50 and $C_r$@100, reiterating the improvement of these components.
\begin{table}
\centering
\small
\begin{tabular}{c|ccc}
\toprule
 & $C_r$@10 & $C_r$@50 & $C_r$@100 \\ \hline
PR & 0.034 & 0.202 & 0.452 \\
SEISA & 0.028 & 0.210 & 0.486 \\
EBM & 0.043 & 0.219 & 0.446 \\
PUL & \textbf{0.026} & 0.181 & 0.424 \\\hline
MOOC & 0.028 & 0.166 & 0.437 \\
MOOC-C & 0.028 & 0.162 & 0.412 \\
MOOC-CG & 0.028 & \textbf{0.160} & \textbf{0.386} \\ \bottomrule
\end{tabular}
\caption{Online evaluation results.}
\label{ACR}
\end{table}

\section{Related Works}

Our work is based on phrase extraction in MOOCs \cite{pan2017course} and is relevant to the set expansion problem, which takes a set of seed entities as input to expand a single category.

Google Sets was an early set expansion system. It spawned quite a few set expansion techniques, such as Bayesian Sets \cite{ghahramani2006bayesian}, SEAL \cite{wang2007language}, SEISA \cite{he2011seisa} and others \cite{sarmento2007more,wang2008iterative,wang2015concept}. They mainly leverage the similarity between entities measured by their co-occurrences in web texts, wrappers and lists. For example, SEISA employs iterative similarity aggregation and SEAL employs PageRank. Recently, SetExpan \cite{shen2017setexpan} extend previous works by selecting context features and \cite{mamou2018term} skillfully employ five different type of context information and gain a very competitive result.

Distinctively, PU-Learning methods \cite{li2010distributional,wang2017sentiment} transform set expansion into a two-class classification problem. A seed set is regarded as a set $P$ of positive examples and candidate set is a set $U$ containing hidden positive and negative cases. The task of filtering the candidate set turns to building a classifier to test if each candidate member is positive or not and this inspires our classification work.

Our approach also benefits from theories of pedagogy. Concept space was first proposed to benefit knowledge comprehension in education\cite{hori1997concept}, and was gradually employed in domain ontology representation; Online games were already used in \cite{kiili2005digital,threatt2014game} education and were also applied for crowdsourcing information collection\cite{yang2018cost}. Both of them significantly affected our design of model.

\section{Conclusions and Future Work}
We conduct a new investigation on automatically course concept expansion in MOOCs.
We precisely define the problem and propose an active model to search external knowledge base for candidate concepts and detect high-quality ones with a classifier.
Moreover, we design a game-based mechanism to subtly involve human efforts in model optimization.
Experimental results on online courses with different domains validate the effectiveness of the proposed method. Promising future directions would be to investigate how to utilize user interaction in MOOCs more adequately, as well as how attributes of course concepts can help expanding.

\subsection*{Acknowledgments}

The work is supported by NSFC key project (U1736204, 61533018, 61661146007), Ministry of Education and China Mobile Joint Fund (MCM20170301), and THUNUS NExT Co-Lab. 

Zhiyuan Liu is supported by the National Key Research and Development Program of China (No. 2018YFB1004503).

\bibliography{acl2019}
\bibliographystyle{acl_natbib}

\appendix

\section{Probe into the different performance across datasets}
\label{sec:appendix}

In our overall evaluation part, methods' performance in different datasets is undulating: 1) English datasets are tend to provide better performance. 2) ZH-DSA is a roadblock of methods, i.e., each method meet a decline in ZH-DSA. To explore the cause of these phenomenons, we take a further observation on the situation of datasets in two aspects.

\noindent \textbf{Looseness of Course concepts.}

\begin{figure}[h]
    \centering
    \includegraphics[width=0.9\linewidth]{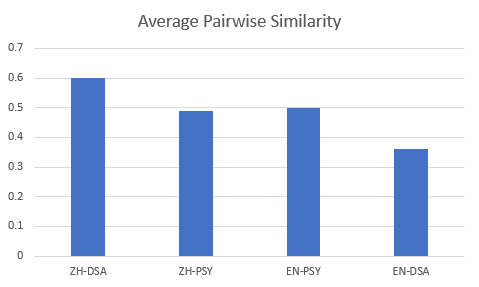}
    \caption{Average pairwise similarity of different datasets.}
    \label{aps}
\end{figure}

We calculate the average pairwise similarity of each dataset, which reflect the alienation of course concepts. The results are shown in Figure \ref{aps}. Obviously the datasets in Chinses contains a more loose $\mathcal{M}$ than English datasets. Thus, when expanding concepts in Chinese courses, the new found concepts are easier to be admitted, for the radius of cluster may be too large to intercept low-quality concepts. In another word, semantic drift is more prone to happen in the two Chinese datasets. Therefore, it is necessary to avoid semantic drifts in real concept expansion of MOOCs.

\noindent \textbf{Specificity of Samples in test set.} 

Despite above observation, we still cannot explain the performance decline in ZH-DSA, for ZH-PSY also provide a competitive result. A deeper investigation is formed by finding out the sample distribution of two Chinese classification results. We reduce the feature of all test concepts to a 2-Dimension vector and differentiate their colors according to actual classification results. When comparing the samples in the two test sets, we can obtain two main observations. For one thing ZH-PSY contains more positive samples. For another thing, the positive and negative samples are more blended in ZH-DSA. The fundamental cause of these characteristics may be the nature of the courses. The course of \emph{Data Structure and Algorithm} in Chinese is interdisciplinary of Computer Science and Mathematics while `Intrduction to Psychology' in Chinese is much more `pure', only containing one domain. Thus, heterogeneous information is indispensable to some extent, for it can effectively utilize different features to query proper expanded concepts.

\begin{figure}
\centering
\subfigure[Sample distribution of ZH-DSA.] { \label{fig:a1}
\includegraphics[width=0.80\linewidth]{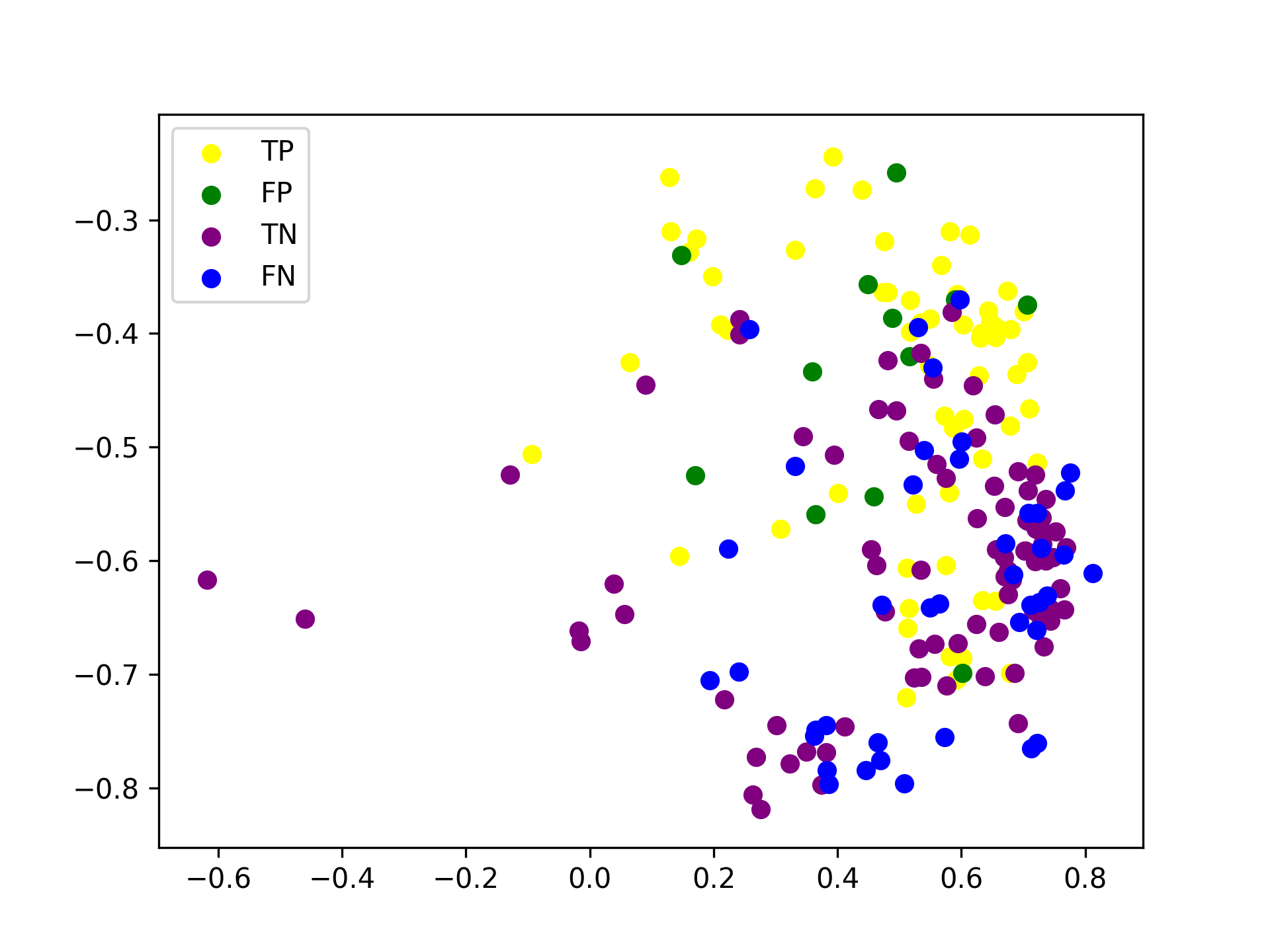}
}
\subfigure[Sample distribution of ZH-PSY.] { \label{fig:b1}
\includegraphics[width=0.80\linewidth]{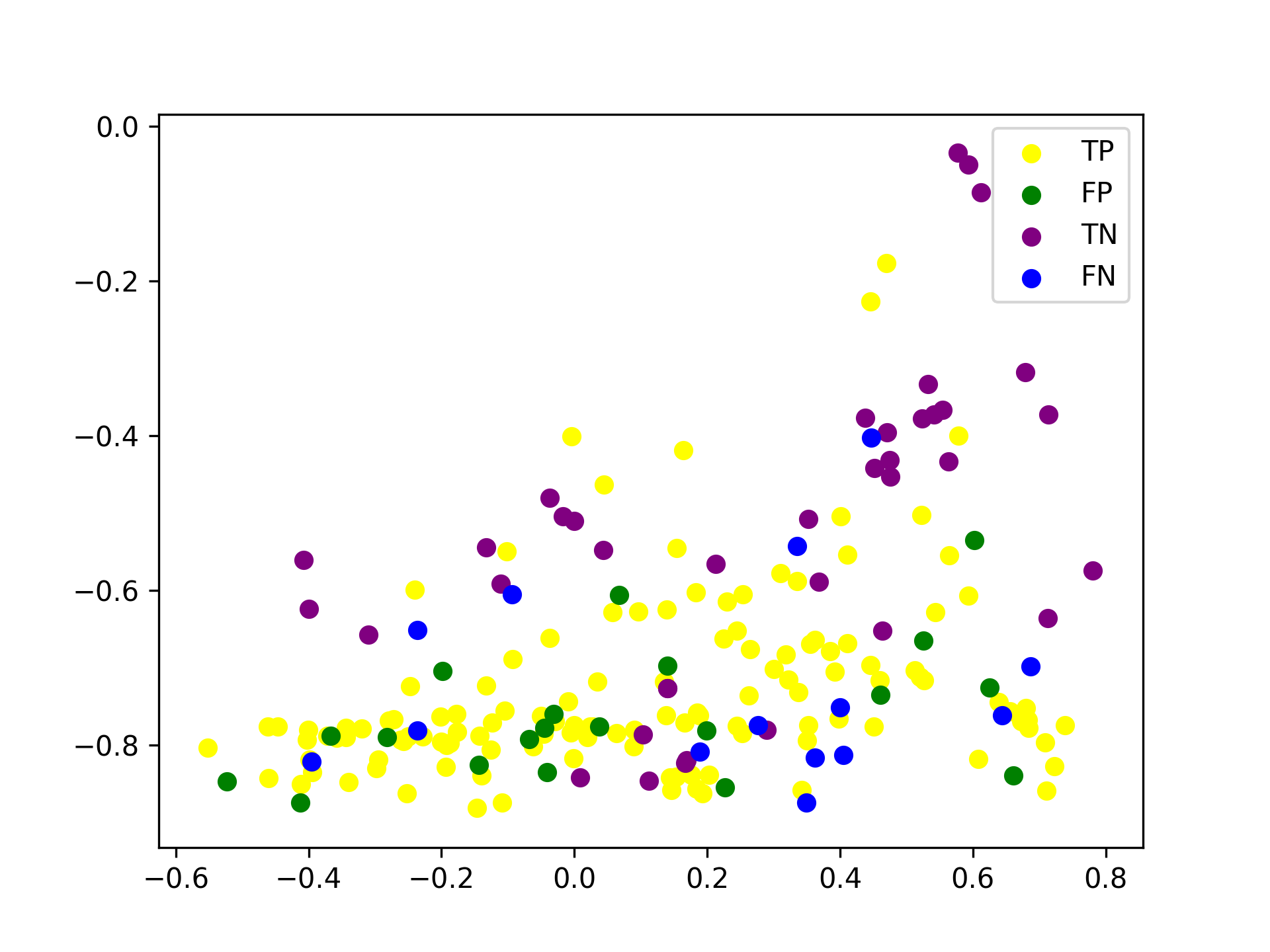}
}
\caption{The sample distribution of two Chinese datasets. TP, FP, TN, FN are respectively \emph{True Positive}, \emph{False Positive}, \emph{True Negative}, \emph{False Negative} samples.}
\label{fig2}
\end{figure}

\end{document}